# Urdu Word Segmentation using Conditional Random Fields (CRFs)

**Haris Bin Zia**
ITU, Pakistan
haris.zia@itu.edu.pk

**Agha Ali Raza**
ITU, Pakistan
agha.ali.raza@itu.edu.pk

**Awais Athar**
EMBL-EBI, UK
awais@ebi.ac.uk

## Abstract

State-of-the-art Natural Language Processing algorithms rely heavily on efficient word segmentation. Urdu is amongst languages for which word segmentation is a complex task as it exhibits space omission as well as space insertion issues. This is partly due to the Arabic script which although cursive in nature, consists of characters that have inherent joining and non-joining attributes regardless of word boundary. This paper presents a word segmentation system for Urdu which uses a Conditional Random Field sequence modeler with orthographic, linguistic and morphological features. Our proposed model automatically learns to predict white space as word boundary as well as Zero Width Non-Joiner (ZWNJ) as sub-word boundary. Using a manually annotated corpus, our model achieves $F_1$ score of 0.97 for word boundary identification and 0.85 for sub-word boundary identification tasks. We have made our code and corpus publicly available to make our results reproducible.

## Title and Abstract in Urdu

اردو میں تقطیع الالفاظ بذریعہ کنڈیشنل رینڈم فیلڈز

جدید ترین نیچرل لینگویج پراسیسنگ کے الگورتھم رواں تحریر کی الفاظ میں تقسیم پر بہت انحصار کرتے ہیں۔ اردو ان زبانوں میں شامل ہے جن میں خودکار طریقے سے یہ تقطیع الالفاظ ایک پیچیدہ عمل ہے کیونکہ اردو لکھتے وقت الفاظ کے درمیان خالی جگہ اکثر حذف کر دی جاتی ہے اور کبھی غیرضروری طور پر داخل بھی کر دی جاتی ہے۔ یہ مسئلہ جزوی طور پر عربی رسم الخط کے استعمال کی وجہ سے ہے جس میں حدودِ الفاظ سے قطع نظر، حروف کی متصلہ اور غیر متصلہ، دونوں اشکال پائی جاتی ہیں ۔ اس مقالہ میں ہم اردو کے لیے ایک خود کار نظام پیش کر رہے ہیں جو کسی بھی اردو تحریر کی بآسانی تقطیع کر سکتا ہے۔ یہ نظام املائی، لسانی اور معنوی اوصاف کا استعمال کرنے والے ایک کنڈیشنل رینڈم فیلڈ ماڈل پر مبنی ہے۔ ہمارا مجوزہ ماڈل خود کار طریقے سے خالی جگہ کی لفظی حد کے طور پر پیشگوئی کرنا سیکھتا ہے اور اس کے ساتھ زیرو وڈتھ نان جائنر (ZWNJ) کی ذیلی لفظ کی حد کے طور پر بھی پیشن گوئی کرتا ہے۔ ایک دستی نشان شدہ کارپس کا استعمال کرتے ہوئے، ہمارے ماڈل کو لفظی حد کی شناخت کے لیے 0.97 کا اور ذیلی لفظ کی حد کی شناخت کے لیے 0.85 کا $F_1$ سکور حاصل ہوتا ہے۔ اپنے نتائج کو قابل تکرار بنانے کے لیے ہم نے اپنے کوڈ اور کارپس کو عمومی طور پر دستیاب کر دیا ہے۔

## 1 Introduction

Word segmentation is an essential step for most Natural Language Processing (NLP) tasks. Most of the state-of-the-art NLP applications such as Machine Translation, Parts-of-Speech (POS) tagging etc. operate at word level. English and other languages that use Latin script usually rely on white spaces to mark word boundaries, with some exceptions like compound words, and exhibit less segmentation issues than Asian languages like Mandarin, Lao, and Urdu etc. that do not have consistent word boundary markings. In the absence of word boundary markers, tokenization of text for further processing is a non-trivial and challenging task.

Urdu, the official language of Pakistan, is an Indo-Aryan language with over 163 million speakers[1] worldwide. It uses Arabic script with segmental writing system. More specifically it uses an *abjad* system where consonants and (most) long vowels are necessarily written while short vowels (diacritics) are

---

[1] https://www.ethnologue.com/language/urd

optional. Urdu is bidirectional and characters are written from right-to-left while numerals are written from left-to-right. A sentence written in Urdu along with its English translation is given below:

<div dir="rtl">اردو پاکستان کی قومی زبان ہے۔</div>

Urdu is the national language of Pakistan.

When Urdu script is written in digital form, white space is not used for word boundary alone but also serves as a sub-word boundary marker as discussed in subsequent sections. Due to this absence of a clear word boundary marker, Urdu exhibits complex segmentation issues for natural language processing as well as information retrieval. In this paper, we present a system to solve this problem of word tokenization. **The major novel contributions of this paper are:**
1. A publicly available annotated corpus for Urdu word segmentation.
2. A Conditional Random Field (CRF) word segmentation utility for Urdu based on linguistic and orthographic features.

The remainder of this paper is structured as follows: Section 2 reviews segmentation techniques. We then present Urdu orthography and writing system in Section 3. Section 4 briefly discusses challenges in Urdu word segmentation. We present our corpus and model in section 5 and conclude in section 6.

## 2 Literature Review

Several techniques have been applied to solve segmentation issues for different world languages. Wong and Chan (1996) proposed a rule-based maximum-matching (max-match) dictionary look-up for Chinese word segmentation. This technique does not take into account the context of words and also never generates short valid words. Another rule-based variant of max-match exists that first generates all possible segmentations of character sequences and then selects the best one based on heuristics like minimum length words etc. (Sornlertlamvanich, 1993; Ping and Yu-Hang, 1994; Nie et al., 1994).

Statistical methods like n-grams have also been extensively used to segment character sequences into words and have proven very effective (Kawtrakul, 1997; Aroonmanakun, 2002). However, performance of statistical word segmentation methods relies heavily on the quality of training corpora and computationally expensive higher order n-grams to capture long distance dependencies. Charoenpornsawat et al. (1998) use context-based features like Winnow (Blum, 1997) for Thai word boundary identification.

Huihsin et al. (2005) proposed a conditional random field (CRF) sequence modeler for Chinese word segmentation. Their feature set consisted of simple n-gram features e.g. unigram and bigram features. Monroe et al. (2014) used the same CRF sequence modeler but with extended linguistic features for Arabic word boundary identification. They have reported $F_1$ score of 0.92 on Egyptian Arabic Treebank[2] and 0.98 on Arabic Treebank[3]. Our proposed model is similar to that of Monroe et al. (2014) but based on a different feature set.

More recently Cai and Zhao (2016) and Cai et al. (2017) investigated the use of neural language models with word and character-based embedding for efficient Chinese word segmentation and achieved better results than traditional segmentation algorithms. As with all deep-learning architectures, their model is data hungry and needs much training data, which is a barrier for low-resource languages.

There has been some research for Urdu word segmentation (Durrani and Hussain, 2010) using hybrid techniques such as rule-based methods relying on max-match, statistical methods such as n-grams together with the POS tags of words. Their best technique have achieved error detection rate of 85.8% for space insertion and space omission errors.

To our knowledge, there is no existing CRF based model for Urdu word segmentation along with a publicly available corpus and gold standard that can act as a reproducible reference for this task.

## 3 Urdu Writing System

Urdu is written using Arabic script in a cursive format (Nastaliq style) from right to left using an extended Arabic character set. The character set includes 37 basic and 4 secondary letters, seven diacritics, punctuation marks and special symbols (Hussain & Afzal, 2001; Afzal & Hussain, 2001; Hussain, 2004).

---
[2] LDC2012E{93,98,89,99,107,125}, LDC2013E{12,21}
[3] LDC2010T13, LDC2011T09, LDC2010T08

When characters in Urdu character-set join to form words, they can acquire different shapes. Based on context, a character may have up to four shape variants: 1) word initial 2) word medial 3) word final, and 4) isolated. Characters that can acquire all of the four shapes are known as *joiners* while one that only have two forms i.e. final and isolated are termed as *non-joiners*. Urdu joiners and non-joiners are shown in Table 1.

| Joiners | ب پ ت ٹ ث ج چ ح خ س ش ص ض ط ظ ع غ ف ق ک گ ل م ن ہ ی |
|---|---|
| Non-joiners | ا د ڈ ذ ر ڑ ز ژ و ء ے |

Table 1: Urdu joiners and non-joiners.

Inherently, Urdu does not have the concept of white space as a word boundary marker. During digital transcriptions, native Urdu typists use space to get accurate shape of characters instead of using it to mark word boundaries. For example, Urdu writers may insert a space within a word دولت‌مند (rich) to make it visually correct, where the character · represents the ASCII space character. Omitting the space would lead to an incorrect visual form, دولتمند, for the same word. On the other hand, writers may omit a space between two separate words like اردوشاعر (Urdu poet) because the shape of characters with or without space remains identical. Due to ambiguous use of white-space in Urdu, it cannot be used as a reliable word boundary marker for NLP applications.

## 4 Challenges of Urdu Word Segmentation

Urdu word segmentation exhibits space omission as well as space insertion challenges. These challenges are discussed in detail by Durrani and Hussain (2010) and are summarized below:

### 4.1 Space Omission

Urdu words ending with non-joiner characters exhibit correct shape even without space and thus the writer may omit a space between words ending with non-joiner characters. Omission of space does not affect readability of words but raises a computation issue. An example is illustrated in Table 2.

| a | اسد·قافلے·کے·صدر·کے·طور·پر·گیا۔ |
|---|---|
| b | اسدقافلے کے صدرکے طورپرگیا۔ |
| c | Asad went as the leader of caravan. |

Table 2: Urdu sentence with all words ending in non-joiner characters (a) with space (shown by a ·) after each word (b) with no space after any word (c) English translation. Computationally, this sentence, which has eight tokens originally (with spaces), reduced to one token (without spaces).

### 4.2 Space Insertion

Another challenge of word segmentation in Urdu arises when two (or more) morphemes join to form a single word. If the first morpheme ends in a joiner character, writers may insert white space to prevent it joining with the next morpheme so that the word retains a valid visual form. If the first morpheme ends in a non-joiner, writers may (correctly) omit the space, as the shape of the word remains the same regardless. The error in this case can be avoided by using the Zero Width Non-Joiner (ZWNJ)[4] Unicode character but Urdu users are generally not aware of its existence and most Urdu keyboards also do not have a direct key mapping for this character. Thus an extra space within a word creates two (or more) separate tokens for a single word and creates a computational problem. The space insertion problem exists in the following cases:

- Affixation: to keep affixes separate from stem.
- Compounding: to keep words compounded together from visually merging.
- Reduplication: to keep reduplicated words from combining.
- Foreign word: to enhance readability of transliterated words.
- Abbreviation: to keep letters separate when foreign abbreviations are transliterated.

---
[4] The standard ISO symbol for ZWNJ is ⎯ which has been approximated by ‌ in this paper to avoid formatting issues

| Case | a | b | c | Translation |
|---|---|---|---|---|
| Affixation | خوش۔نصیب | خوشنصیب | خوش‌نصیب | Lucky |
| Compounding | نظم۔و۔ضبط | نظموضبط | نظم‌و‌ضبط | Discipline |
| Reduplication | دہوم۔دہام | دہومدہام | دہوم‌دہام | Pomp & Show |
| Foreign word | فٹ۔بال | فٹبال | فٹ‌بال | Football |
| Abbreviation | پی۔ایچ۔ڈی | پیایچڈی | پی‌ایچ‌ڈی | Ph.D. |

Table 3: Example of space insertion (a) incorrect multiple tokens with space (·), correct shape (b) single token, incorrect shape (c) single token with ZWNJ (‌) as sub-word boundary, correct shape.

## 5 Urdu Word Segmentation Model

In order to accurately and efficiently solve space omission and insertion issues, we propose a Conditional Random Field (CRF) sequence model that uses linguistic and orthographic features to predict white space as word boundary and ZWNJ as sub-word boundary markers. Our model takes as input a concatenated sequence of characters and outputs sequence of words with space as word boundary and ZWNJ as sub-word boundary markers as shown in Figure 1. A description of our data annotation, experiments, and results is as follows.

### 5.1 Data Annotation

Unlike resource-rich languages such as English and Arabic that have abundant publicly accessible linguistic resources, Urdu is relatively under-resourced and does not have any segmentation benchmark corpus. To overcome this shortcoming, we manually annotated a corpus of approximately 111,000 tokens by using the Urdu-Nepali-English parallel corpus[5] as a starting point. This corpus is a parallel corpus to the common English source from PENN Treebank corpus where the source English sentences were news stories from Wall Street Journal (WSJ). We then used the CLE Urdu corpus cleaning application[6] to mark white space as word boundary and ZWNJ as sub-word boundary markers as per the rules proposed by Rehman et al. (2011). A summary of these rules is being reproduced here for the convenience of the reader:

- White space after each word ending.

---
[5] http://www.cle.org.pk/software/ling_resources/UrduNepaliEnglishParallelCorpus.htm
[6] http://www.cle.org.pk/software/langproc/corpuscleaningH.htm

- ZWNJ between two roots or stems of X-Y compounding e.g. انشاءاللہ

- ZWNJ between two roots or stems of X-e-Y compounding e.g. وزیراعظم

- ZWNJ before and after و in X-o-Y compounding e.g. نظم‌و‌ضبط

- ZWNJ between reduplicated words e.g. روٹی‌وٹی

- ZWNJ between prefix and root in case of prefixation e.g. خوش‌اخلاق

- ZWNJ between root and suffix in case of suffixation e.g. شادی‌شدہ

- ZWNJ between multiple morphemes of a single transliterated word e.g. فٹ‌بال

- ZWNJ between multiple morphemes of a transliterated abbreviation e.g. پی‌ایچ‌ڈی

The complete corpus was annotated by the first author. To calculate the inter-annotator agreement, 100 sentences (21,781 characters) were annotated by one of the co-authors and the inter-annotator agreement was found to be 0.98 as measured using Cohen's Kappa. Both annotators are native Urdu speakers with a background of computer science and NLP, and use Urdu scripts in both printed and digital form on a regular day-to-day basis.

Next, we applied Unicode text normalization to convert multiple equivalent representations of characters to their consistent underlying normal forms and removed all diacritics and multiple spaces from the text. Some of these characters, such as diacritics, are good indicators of word boundaries; however, it is a common practice of first language writers of Urdu to exclude diacritics in all informal and most formal forms of writing. The reader simply uses contextual information to interpret the correct word. Therefore, to model the real-life situation more closely we decided to omit all diacritics from our corpus. Omission of diacritics makes the segmentation task harder (as their removal loses a major word disambiguation and hence possibly word segmentation cue) but makes it unbiased at the same time. For the sake of comparison, we report results with diacritics present as well as removed in later sections.

The corpus consists of 4,325 sentences which cover most Urdu morphological and graphical constructs such as affixes, compounds, reduplicates, foreign words (transliterated from Latin to Arabic script) as well as abbreviations (transliterated from Latin to Arabic script) and words ending in both joiner and non-joiner characters. We have made this corpus publicly available on GitHub[7] along with our pipeline. For evaluation purposes, we have split the data in terms of sentences: 3,500 for training (90K tokens) and 825 for testing (21K tokens). The test set contains 20,264 word boundary spaces and 1,200 ZWNJ sub-word boundary markers.

### 5.2 Model

A CRF model, proposed by Lafferty et al. (2001), is defined as $P(Y|X;w)$ where $X = \{x_1,...,x_n\}$ is a sequence of input and $Y = \{y_1,...,y_n\}$ is a predicted sequence of labels. We use a linear-chain model (Green and DeNero, 2012) with $X$ as concatenated sequence of characters and $\hat{Y}$ is chosen as per the following decision rule.

$$\hat{Y} = \text{argmax} \sum w^T \phi(X, y_i); i = 1,...,n$$

Where $\phi$ is a feature map defined in section 5.3. Our model classifies each $y_i$ as one of the following:

- I: continuation of a word or sub-word.

- $B_W$: beginning of a word.

- $B_S$: beginning of a sub-word.

---
[7] https://github.com/harisbinzia/Urdu-Word-Segmentation

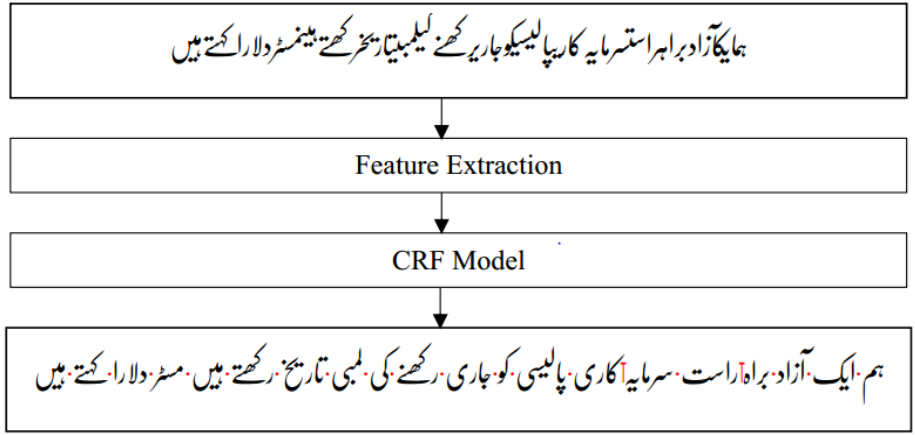

Figure 1: Urdu word segmentation pipeline. The input is a concatenated sequence of characters and the output is a sequence of words with white space (·) as word boundary marker and ZWNJ (|) as sub-word boundary marker. In this example براہراست (live) is a compound word that consists of two morphemes براہ and راست so our model predicted ZWNJ between them. Similarly, سرمایہ‌کاری (investment) is an example of affixation with ZWNJ between root سرمایہ and suffix کاری.

### 5.3 Feature Engineering

We experimented with different orthographic and linguistic features such as character windows of varying lengths, joining/ non-joining behavior etc. and selected the best ones for our model. Table 4 summarizes the effect of each feature on the performance of the model. Our final CRF model employs the following features:

- N-grams consisting of the current character and up to three preceding and three succeeding characters e.g. for each $i^{th}$ character $x_i$, the character sequence $\{x_{i-3},...,x_i,...,x_{i+3}\}$ is considered.

- Whether the current character is a digit.

- Whether the current character is a joiner.

- Unicode class of current character.

- Direction of current character. Urdu is bidirectional where basic/secondary characters and diacritics are written from right-to-left while numeric characters from left-to-right.

### 5.4 Results

We have used precision, recall, and $F_1$ measure as our evaluation metrics as they provides a more informative assessment of the performance than the word level and character level error rates. On an unseen undiacritized test set of 825 sentences (21K tokens) our model achieved $F_1$ score of 0.97 for word boundary and 0.85 for sub-word boundary identification. The detailed results are shown in Table 5 and 6.

Not surprisingly, the $F_1$ score for sub-word boundary identification is slightly higher for diacritized text as some diacritics are very indicative features of sub-word boundary e.g. ٔ in compounding. Diacritized text also has high precision over undiacritized text for word boundary prediction as the diacritic ٌ is a clear indication of word boundary.

We also report the macro and micro $F_1$-measures. However the results do not show much improvement between diacritized and non diacritized corpora. One possible explanation is that the corpus is very sparsely diacritized with only 5,237 diacritics in 111K tokens. Solution to this problem via automatic diacritization is beyond the scope of the current paper.

|  | $F_1$ | |
|---|---|---|
| Features | Word Boundary (Space) | Sub-word Boundary (ZWNJ) |
| Current character | 0.59 | 0 |
| +1 character window | 0.77 | 0.07 |
| +2 character window | 0.85 | 0.46 |
| +3 character window | 0.92 | 0.72 |
| +isDigit | 0.95 | 0.79 |
| +isJoiner | 0.96 | 0.84 |
| +Unicode class | 0.97 | 0.85 |
| +Direction | 0.97 | 0.85 |

Table 4: Results with different feature set.

a) Undiacritized test set

| Gold \ Predicted | I | $B_W$ | $B_S$ |
|---|---|---|---|
| I | 59,149 | 474 | 42 |
| $B_W$ | 574 | 19,637 | 53 |
| $B_S$ | 119 | 116 | 965 |

b) Diacritized test set

| Gold \ Predicted | I | $B_W$ | $B_S$ |
|---|---|---|---|
| I | 60,254 | 405 | 44 |
| $B_W$ | 560 | 19,660 | 44 |
| $B_S$ | 115 | 88 | 997 |

Table 5: Confusion matrices for word and sub-word boundary identification.

|  | Boundary | Precision | Recall | $F_1$ |  |
|---|---|---|---|---|---|
| **Without Diacritics** | Word | 0.97 | 0.97 | 0.97 | macro-$F_1$= 0.91 |
|  | Sub-word | 0.91 | 0.80 | 0.85 | micro-$F_1$= 0.96 |
| **With Diacritics** | Word | 0.98 | 0.97 | 0.97 | macro-$F_1$= 0.92 |
|  | Sub-word | 0.92 | 0.83 | 0.87 | micro-$F_1$= 0.96 |

Table 6: Test set results with and without diacritics.

## 6 Conclusion & Future work

We present a CRF sequence modeler for word and sub-word boundary identification in Urdu text using orthographic and linguistic features. Our best model achieves $F_1$ score of 0.97 and 0.85 for word and sub-word prediction tasks respectively. We also present handcrafted training and testing data that we have made publicly available to allow for reproducible research.

The presented system is trained using a manually crafted corpus of 4,325 sentences (111K tokens) which is relatively small compared to segmentation benchmark corpora of Arabic and Chinese. As a future direction we will extend our work to tag a much bigger corpus with space as word boundary markers and ZWNJ as sub-word boundary markers. We will explore more linguistic features for the CRF model to increase performance of sub-word boundary identification task. We also plan to explore neural models such as RNN and Bidirectional RNN for Urdu word and sub-word boundary prediction.